\documentclass[runningheads]{llncs}
\usepackage[T1]{fontenc}
\usepackage{algorithm,algpseudocode,physics,graphicx}
\usepackage{pgfplots}
\usepackage{pgfplotstable}
\pgfplotsset{compat=1.7}
\usepackage{tikz}

\begin{document}
\title{Study of the Proper NNUE Dataset}
%
%
\author{Daniel Tan\inst{1}\orcidID{0000-0002-5715-8584} \and
Neftali Watkinson Medina\inst{1}\orcidID{0000-0001-8222-8591}}
\authorrunning{Daniel Tan et al.}
%
\institute{University of California, Riverside CA 92521, USA}
%
%
\maketitle              
\begin{abstract}
NNUE (Efficiently Updatable Neural Networks) has revolutionized chess engine development, with nearly all top engines adopting NNUE models to maintain competitive performance. A key challenge in NNUE training is the creation of high-quality datasets, particularly in complex domains like chess, where tactical and strategic evaluations are essential. However, methods for constructing effective datasets remain poorly understood and under-documented. In this paper, we propose an algorithm for generating and filtering datasets composed of "quiet" positions—positions that are stable and free from tactical volatility. Our approach provides a clear methodology for dataset creation, which can be replicated and generalized across various evaluation functions. Testing demonstrates significant improvements in engine performance, confirming the effectiveness of our method.

{}

\end{abstract}
\section{Introduction}

NNUE (Efficiently Updatable Neural Networks) is a chess evaluation technique that utilizes the incremental update technique to quickly update the model for the next evaluation value. Instead of recalculating the entire sparse neural network matrix multiplication calculation from scratch, one only needs to update the small changes in a position that results from the move to get the next evaluation. Through incremental update, NNUE is able to efficiently calculate evaluations of millions of positions during an alpha beta negamax. Through this technique, NNUE has dramatically increased the elo playing strength of top chess engines over the several months ~\cite{SF_ELO}.\\*

\noindent Despite NNUE’s widespread adoption, the process for creating an effective dataset remains poorly understood and documented. There are vague suggestions or ambiguous descriptions on dataset creation, but no implementation details and no explanation for why those techniques work. In this paper, we aim to address this gap by asking: What makes a good dataset for training NNUE models?

\section{Background}

NNUE (Efficiently Updatable Neural Networks) is a neural network evaluation function introduced by Yu Nasu, first used in the Japanese Shogi engine YaneuraOu ~\cite{NNUE}. Following its success in Shogi, NNUE was adapted for other board games, including Western Chess ~\cite{SF} and Chinese Chess ~\cite{Pikafish}. Since then, it has revolutionized all engines across all different chess variants, with nearly all top engines implementing NNUE models to remain competitive.\\*

\noindent A critical aspect of NNUE’s success lies in the quality of the datasets used to train the network. A well-curated dataset ensures that NNUE can learn accurate evaluations of positions during training. However, creating an effective NNUE dataset remains poorly documented. While ~\cite{SF_DOC} provides an in-depth explanation of SIMD programming for engine evaluation functions, it offers no guidance on how to build a robust training dataset. Similarly, generating good datasets is often left to trial and error, requiring developers to interpret complex source code without clear instructions.\\*

\noindent Some sources, such as ~\cite{NN_for_Chess}, offer guidance on implementing neural network forward propagation but does not teach on how to properly collect and filter training data. For example, collecting random position/score pairs results in suboptimal training. Effective datasets must focus on “quiet” positions—positions that are stable and unlikely to change drastically in material or tactical balance. Noisy positions, which involve potential forks, captures, or tactical threats, should be excluded to prevent poor network performance.\\*

\noindent In this paper, we propose a methodology for generating high-quality training datasets for NNUE, focusing on selecting quiet positions and filtering out noisy or tactically unstable positions.

\section{Definitions}

This is a brief summary of the terminology used throughout this paper. We provide the following definitions to ensure clarity and prevent confusion.

\subsection{Negamax}

The negamax ~\cite{negamax} algorithm is a variant of the minimax search that utilizes the zero sum property of a two player game to simplify the minimax implementation. Instead of using two subroutines for the min and max player, negamax uses just one subroutine to do minimax, and the mathematical formula below is used to switch the evaluation scores between the two different sides.

\begin{equation}
\textit{min(a,b)} = \textit{-max(-a,-b)}
\end{equation}

\subsection{Quiescence Search}

The quiescence ~\cite{quiescence} search is an algorithm used to extend the search at unstable nodes in the game tree. The idea behind quiescence search is to only evaluate quiet positions where there are no winning tactical moves to be made. A crude search algorithm may find that a player wins an extra pawn on X depth, but might lose an entire rook immediately on the next move due to the search stopping at X depth and not exploring further. Using quiescence search to extend the search endpoints mitigates the horizon effect where a search may be unable to see further into a position due to limitations of search.

\subsection{Centipawn}

A unit of measurement of a point advantage in Chess evaluation. A centipawn is one hundredth of a pawn. In this system, a one pawn advantage indicates that a player is at least one pawn up in material advantage.

\subsection{Piece Square Tables}

A piece square table ~\cite{Adaboost} is a basic method of assigning specific pieces to specific squares. A table is created for each piece, and table values are given to each square. The higher the value of a square, the more likely a negamax function is going to place the piece on the particular square.

\subsection{Elo}

The Elo rating system ~\cite{Elo} is the measurement of the relative playing skill levels of players in Chess/Go/other games. Elo is calculated using the following formula:

\begin{equation}
\textit{performance rating} = \frac{\textit{total opponents' rating} + 400 \* (\textit{wins} - \textit{losses})}{\textit{number of games}}
\end{equation}

\noindent To illustrate a few examples of how elo is used, the person with the highest peek Elo rating chess player, according to FIDE (the World Chess Federation i.e. Fédération Internationale des Échecs) ~\cite{FIDE}, is the 2013-2023 World Champion Magnus Carlsen at 2882 peek ~\cite{Magnus_Elo}. The strongest computer chess engine, Stockfish ~\cite{SF}, is rated approximately 3600 Elo ~\cite{CCRL}. With that being said, the strongest computer chess program is significantly stronger than the best human players.

\section{Platform}

When experimenting with what kinds of data works, we implemented a Xiangqi (Chinese Chess) engine. Xiangqi arguably contains many interesting properties that make it better for exploring the power of NNUE compared with Western Chess. In Western Chess, being just a pawn down materially can lead to a loss, but in Xiangqi, the king is so easily exposed to attacks that being several pawns down is not such a problem as long as you have a strong attack ~\cite{kwan_xq}. Xiangqi is not just a game of gaining a material advantage over your opponent; Xiangqi requires players to also judge the coordination, quality, and positioning of the pieces rather than counting up the number of pieces ~\cite{xq_vs_chess,kwan_xq}. As we will discuss in later sections of the paper, NNUE is incredibly effective in Xiangqi.\\*

\noindent Another benefit of Xiangqi over Western Chess is the repetition rules. Unlike Western Chess where perpetual check is a draw, in Xiangqi, a perpetual check or chase is a loss for the player performing it ~\cite{WXF_Rules}. This rule has profound implications. In Western Chess, a player with the advantage needs to spend extra effort making sure the opponent does not use perpetual checks/repetitions to swindle a draw from them; in contrast, a Xiangqi player can ignore perpetual checks or chases and focus on attacking the opponent since the opponent cannot use a repetition to force a draw. As a result, Xiangqi is a much more complex attacking game with much less of a chance to get a draw ~\cite{xq_vs_chess}.\\*

\section{Methodology}

To create a good dataset to test NNUE on, we collected 44000 grandmaster games from ~\cite{DPXQ,WXF_GAMES}. These Xiangqi websites have large databases of various official tournament games between Xiangqi masters from China, Vietnam, and other Asian countries which play Xiangqi from the years 2008 to 2024. From the recorded tournament games, we extracted all the moves and positions from recorded grandmaster games.\\*

\noindent Adding to the existing grandmaster games, we also generated another 30000 synthetic games through self play between different engines ~\cite{eleeye,Pikafish} and our own engine. We took random positions found in the grandmaster games, and generated additional games through self play between different engines. Just like the grandmaster games, we extracted all the moves and positions found in the recorded synthetic games.

\begin{figure}[!ht]
  \centering
  \begin{minipage}[b]{0.45\textwidth}
    \includegraphics[width=\textwidth]{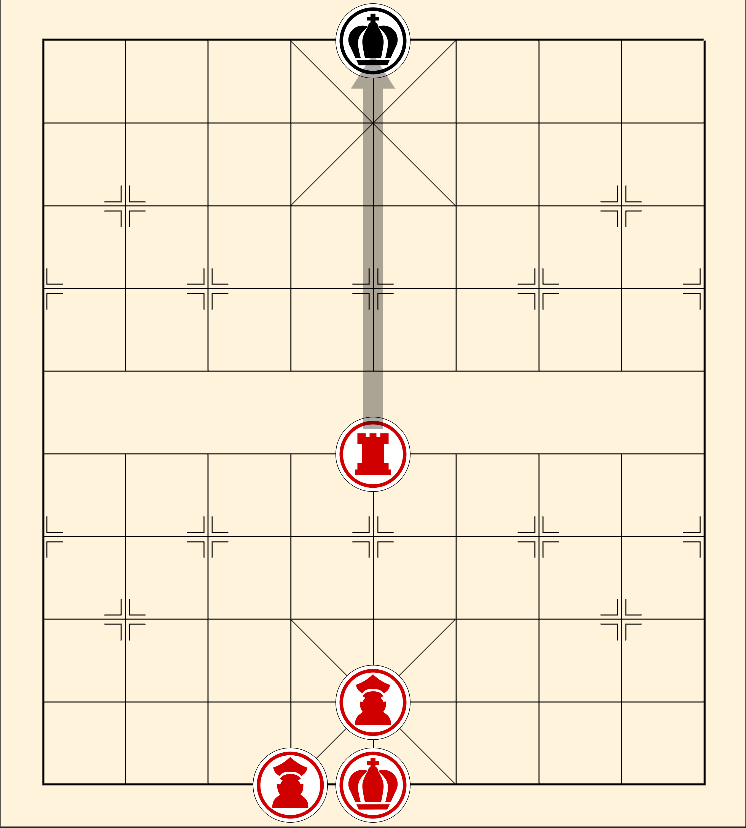}
    \caption{In this position, the rook checks the king. Because of the check, this is not a quiet position to be included in the dataset.\\*\\*}
  \end{minipage} 
  \hfill
  \begin{minipage}[b]{0.45\textwidth}
    \includegraphics[width=\textwidth]{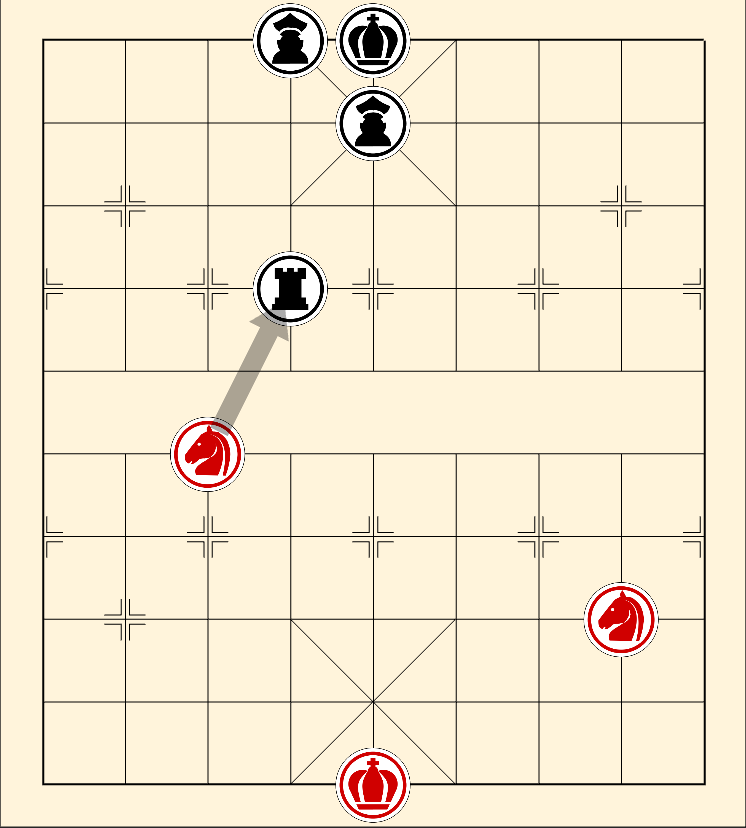}
    \caption{In this position, the knight can capture the rook for free. Because of the free capture of the rook, this is not a quiet position to be included in the dataset.\\*}
  \end{minipage}
  \hfill
  \begin{minipage}[b]{0.45\textwidth}
    \includegraphics[width=\textwidth]{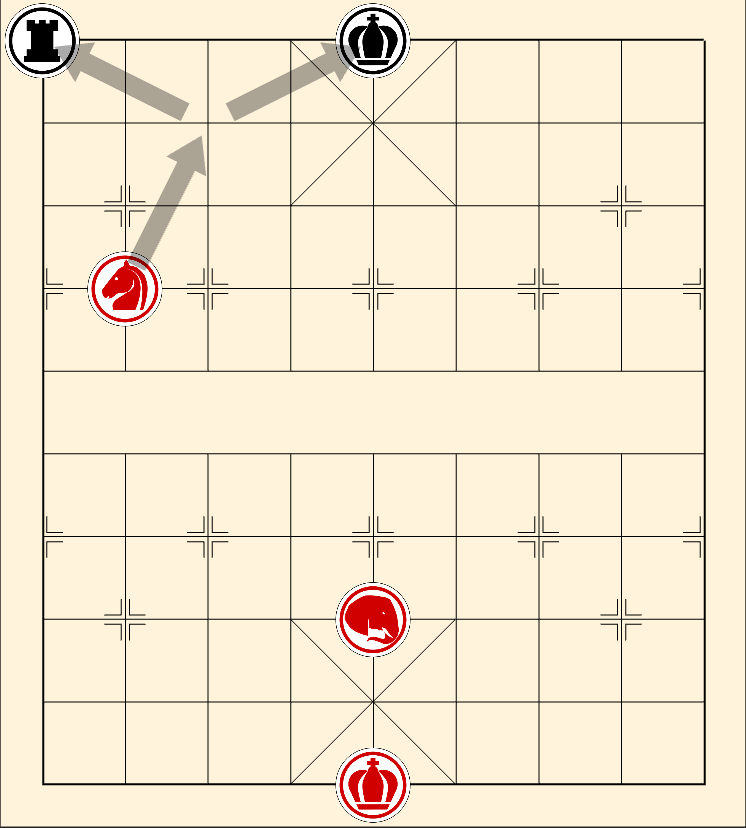}
    \caption{In this position, the knight can fork the king and rook, winning a free rook. Because of the fork, this is not a quiet position to be included in the dataset.\\*}
  \end{minipage}
  \hfill
  \begin{minipage}[b]{0.45\textwidth}
    \includegraphics[width=\textwidth]{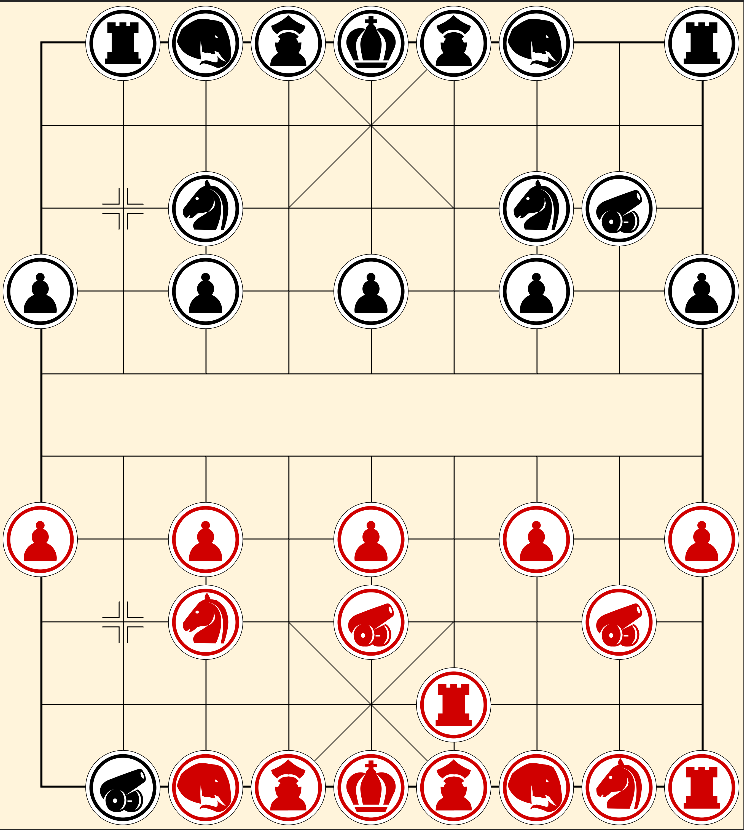}
    \caption{This position contains no significant changes in material imbalance, whether through a fork or tactic, only exchanges of relatively equal pieces. This is a quiet position useful to the dataset.}
  \end{minipage}
\end{figure}

\section{Implementation}

\noindent From the positions extracted out of the grandmaster games and synthetic games, we loop over all the possible moves, and generate the potential training/validation dataset from positions. We filter out noisy or tactically unstable positions and focus on selecting diverse high quality quiet positions for training. We generate a diverse set of position/evaluation pairs from all sorts of different kinds of positions.\\*

\noindent With that being said, using noisy or tactically unstable positions drastically ruins the training of the neural network. Effects of noisy positions include the training algorithm refusing to converge due to the noise, the mean square error loss being drastically larger, and the network taking longer to train for poorer results. In terms of the effect on engine playing style, the engine might randomly sacrifice pieces for no good reason, be unable to calculate material correctly, or hesitate at crucial moments of the game.\\*

\noindent We filter out noisy or unstable positions that would harm the training/validation and find proper quiet positions to include in the dataset. Not all positions are good positions to include in the training/validation dataset. Only quiet positions, positions where subsequent move sequences, such as captures, forks, or checks, do not drastically change the material imbalance are good positions for the dataset. If a rook can be captured for free in the next turn by a pawn in such a way the evaluation score would drop by 500 centipawns, then the position is not quiet. Conversely, if a position contains no possible forcing move sequences that will drastically alter the material imbalance, then it can be considered a quiet position. Good examples of quiet positions include figure 4, while bad positions that should be filtered out are figures 1, 2, and 3.\\*

\noindent A check against the king (see Figure 1) could lead to a checkmate or other tactical advantages. Because of the volatile nature of a check position, we defined it such that quiet positions cannot contain a check against the king. All positions containing a check against a king are filtered out of the dataset. Not all positions without a check can be considered quiet (e.g. a fork of two important pieces), but all quiet positions must be without a check.\\*

\noindent Good datasets for NNUE are generated primarily from quiet positions ~\cite{training_dataset}. To determine whether position $\textit{P}$ is a viable position, find the difference between the static evaluation $\textit{E}$ and the quiescence search evaluation value $\textit{Quiescence(P)}$. If the difference is greater than a certain margin $\textit{M}_1$, then position $P$ is filtered out of the dataset. According to the definition, a quiescence search extends the search endpoints to mitigate the horizon effect. If the static evaluation differs drastically from the quiescence search evaluation value, it indicates that the position has a capture move available that can drastically shift the evaluation value (e.g. see figure 2 where a rook can be lost for free on the next turn). Through testing, we found that $\textit{M}_1 = 60$ centipawns to be the best value for margin $\textit{M}_1$.\\*

\noindent Another constraint that the NNUE quiet position $\textit{P}$ must meet is that the difference between the static evaluation $\textit{E}$ and the negamax evaluation value $\textit{Negamax(P)}$ of the position must not be greater than a certain margin $\textit{M}_2$. If the difference is greater than a certain margin $\textit{M}_2$, then position $P$ is filtered out of the dataset. If the static evaluation varies drastically from the negamax search evaluation value, it indicates that the position winning due to a strong mating attack or a forcing tactical sequence that will upset the material imbalance dramatically. Examples of positions that this filters out include a knight forking the king and rook and capturing a rook for free in the next turn (see figure 3). We found $\textit{M}_2 = 70$ to be the best value for the margin $\textit{M}_2$.\\*

\begin{algorithm}[!ht]
\begin{algorithmic}[1]
\caption{compute\_dataset}
\Procedure{compute\_dataset}{$G$} 

\State $Dataset \gets \emptyset$
\ForAll {$P \in G$} \Comment{Loop over all positions $P$ in games $G$}

\ForAll {$move \in $ \Call{all\_moves}{$P$}}
  \State \Call{do\_move}{$P$}
  \If {\Call{is\_quiet\_position}{$P$}}
    \State \Call{add\_position\_to\_dataset}{$Dataset$,$P$}
  \EndIf
  \State \Call{undo\_move}{$P$}
\EndFor

\EndFor
\EndProcedure
\\
\Procedure{is\_quiet\_position}{$P$}
\If {\Call{in\_check}{$P$}}
  \State \Return $False$
\EndIf
\State $E \gets$ \Call{evaluate}{$P$}
\If{$\abs{E - \textit{Quiescence(P)}} > M_1$ \textbf{or} $\abs{E - \textit{Negamax(P)} } > M_2$ }
  \State \Return $False$
\EndIf

\State\Return $True$
\EndProcedure

\end{algorithmic}
\end{algorithm}

\noindent When defining a NNUE dataset, a diversity of positions is necessary for a neural network to fully understand the game. If a certain set of positions is missing from the dataset, then the neural network will fail to correctly recognize the positions and engine play will dramatically suffer. There are many different kinds of positions, such as where one side is overwhelmingly winning in terms of material advantage, positions where both sides are around equally matched, positions in the middlegame, positions in the endgame, open highly tactical positions, and closed positions.\\*

\noindent To illustrate the importance of diversity of positions, if a dataset lacks positions with significant material disadvantage and had only positions where positions are always equal, an engine would throw away 2 rooks for no reason yet believe the positions is equal in material. Positions with a severe material disadvantage need to be marked with serious penalties to tell the engine to avoid those positions.\\*

\noindent A good dataset should have 50\% of the data evaluation points have position evaluation values relative to the side to move, and 50\% of the points have negative evaluation values relative to the side to move. At least 50\% of the data should have evaluation values between -100 and 100, and at least 40\% of the data should contain materially imbalanced positions (i.e. values less than -100 or greater than 100).\\* 

\noindent For the evaluation function used as a starting point, we used the handcrafted evaluation functions ~\cite{Adaboost}. ~\cite{Adaboost} is a simplistic evaluation function that only involves the use of piece squared tables. By using only a simplistic evaluation function, this methodology can be easily to replicated across different machines by copying the piece square table values.\\*

\noindent For the neural network model architecture, we used a simple two layer fully connected neural network with 3240 inputs, 1620 inputs per each player. The hidden layer has 256 inputs. The total amount of megabytes for the neural network is approximately 0.6 MB, making this an incredibly small neural network.

\begin{verbatim}
class NNUE(torch.nn.Module):


    def __init__(self):
        super(NNUE, self).__init__()
        self.feature = torch.nn.Linear(1620, 128)
        self.output  = torch.nn.Linear(256, 1)


    def forward(self, white, black):
        white = self.feature(white)
        black = self.feature(black)
        accum = torch.clamp(torch.cat([white, black], dim=1), 0.0, 1.0)
        return torch.sigmoid(self.output(accum))
\end{verbatim}






\section{Results}






The simple Xiangqi NNUE evaluation function trained from datapoints derived ~\cite{Adaboost} led to a massive increase in strategic understanding and positional play. Playing against the previous handcrafted evaluation function, the NNUE evaluation function was around +100 elo rating playing strength stronger. The upgraded NNUE engine won 65\% of the games against the previous handcrafted evaluation version of the engine. Further training and search improvements improved the engine rating even further.\\*

\noindent To make sure our algorithm works across different evaluation functions, we also used the ~\cite{eleeye} evaluation function to train the dataset, a much more complex handcrafted evaluation function that takes into account piece mobility, trapped pieces, and defense fortresses. The algorithm works perfectly fine for simple piece square tables such as ~\cite{Adaboost} as well as complex evaluations such as ~\cite{eleeye,Pikafish}. The complexity of the evaluation criteria does not change the algorithm in any substantial way.\\*


\noindent In terms of playing style, the previous engine using simplistic piece square tables lacked understanding of piece coordination. Due to the simplistic nature of a piece square table, the previous engine would regularly fall for basic opening traps such as sacrificing four pawns to gain an effective mating attack ~\cite{kwan_xq}. It would take an extremely talented programmer to create a complex handcrafted heuristic to detect such a trap. The new version of the engine using the NNUE evaluation, by contrast, has no such blindspots.\\*

\begin{tikzpicture}
\begin{axis}[
    title=Engine Elo Rating over Months after adopting NNUE,
	xlabel=Months after adopting NNUE,
	ylabel=Elo Rating,
	width=10cm,height=7cm,
    legend style={at={(0.0,.91)},anchor=west}
    ]

\addplot[color=blue,mark=x] coordinates {
    (0, 2400)
	(1, 2500)
	(2, 2530)
	(3, 2550)
	(4, 2595)
	(5, 2630)
	(6, 2660)
};
\end{axis}
\end{tikzpicture}

\noindent Just by switching from a simple piece square table to NNUE, the engine has much better understanding of piece mobility, trapped pieces, tactical understanding, and is generally much more creative than a simple piece square table where one just places a piece on a square just to increase a point value. The engine no longer played like a robot moving pieces to squares with good piece square table values, but rather developed a dynamic AlphaZero playing style based on a human-like intuition.

\section{Future Work}

This algorithm contains most of what is needed to generate a good NNUE training dataset. To reiterate the Results section, we tested our methodology to make sure it worked across many different evaluation functions. Despite this however, this algorithm is not mathematically proven to generate good datasets and may need to be modified to accommodate the nuances of vastly different kinds of evaluation functions. Ultimately, the creation of good training datasets is measured on whether the trained model increases the playing strength of the engine.\\*

\noindent Future work on this will focus on testing many other different evaluation functions as well as different chess variants such as Western Chess, Shogi, Jangqi, or Thai Chess to confirm that this algorithm works independently across different kinds of chess variants and different evaluation functions.

\section{Conclusion}

Despite the advancement of NNUE, the creation of a good dataset is poorly understood and poorly documented. Good dataset creation is a critical part of NNUE training, yet no papers or documentation has touched this topic. Understanding good dataset creation is a matter of tons of guesswork. In this paper, we documented and defined an algorithm for finding defining a good dataset for NNUE. The algorithm can be easily replicated, and produces massive increases in playing strength. The NNUE evaluation function developed a dynamic AlphaZero playing style based on human-like intuition.

%
%
%
\bibliographystyle{splncs04}
%

\end{document}